\documentclass[letterpaper]{article} % DO NOT CHANGE THIS
\usepackage{aaai23}  % DO NOT CHANGE THIS
\usepackage{times}  % DO NOT CHANGE THIS
\usepackage{helvet}  % DO NOT CHANGE THIS
\usepackage{courier}  % DO NOT CHANGE THIS
\usepackage[hyphens]{url}  % DO NOT CHANGE THIS
\usepackage{graphicx} % DO NOT CHANGE THIS
\usepackage{natbib}  % DO NOT CHANGE THIS AND DO NOT ADD ANY OPTIONS TO IT
\usepackage{caption} % DO NOT CHANGE THIS AND DO NOT ADD ANY OPTIONS TO IT
\usepackage{algorithm}
\usepackage{algorithmic}
\usepackage{multirow}
\usepackage{makecell}
\usepackage{algorithm}
\usepackage{algorithmic}
\usepackage{newfloat}
\usepackage{listings}
\usepackage{bibentry}
\usepackage{color}
\usepackage{amsmath}  
\usepackage{amsfonts}

\usepackage{newfloat}
\usepackage{listings}
\DeclareCaptionStyle{ruled}{labelfont=normalfont,labelsep=colon,strut=off} % DO NOT CHANGE THIS
\floatstyle{ruled}
\newfloat{listing}{tb}{lst}{}
\floatname{listing}{Listing}
\title{Attention-based Depth Distillation with 3D-Aware Positional Encoding \\ for Monocular 3D Object Detection}
\author{
    Zizhang Wu\equalcontrib \textsuperscript{\rm 1},
    Yunzhe Wu\equalcontrib \textsuperscript{\rm 1},
    Jian Pu\textsuperscript{\rm 2},
    Xianzhi Li\textsuperscript{\rm 3},
    Xiaoquan Wang\textsuperscript{\rm 1},
}
\affiliations{
    \textsuperscript{\rm 1} ZongmuTech,\\
    \textsuperscript{\rm 2} Institute of Science and Technology for Brain-Inspired Intelligence, Fudan University,\\
    \textsuperscript{\rm 3} Huazhong University of Science and Technology,\\
    \{zizhang.wu, nelson.wu, rockywin.wang\}@zongmutech.com, \\
    jianpu@fudan.edu.cn, xzli@hust.edu.cn
}

\usepackage{bibentry}
\begin{document}

\maketitle
\begin{abstract}

Monocular 3D object detection is a low-cost but challenging task, as it requires generating accurate 3D localization solely from a single image input. Recent developed depth-assisted methods show promising results by using explicit depth maps as intermediate features, which are either precomputed by monocular depth estimation networks or jointly evaluated with 3D object detection. However, inevitable errors from estimated depth priors may lead to misaligned semantic information and 3D localization, hence resulting in feature smearing and suboptimal predictions. To mitigate this issue, we propose $\mathbf{ADD}$, an $\mathbf{A}$ttention-based $\mathbf{D}$epth knowledge $\mathbf{D}$istillation framework with 3D-aware positional encoding. Unlike previous knowledge distillation frameworks that adopt stereo- or LiDAR-based teachers, we build up our teacher with identical architecture as the student but with extra ground-truth depth as input. Credit to our teacher design, our framework is seamless, domain-gap free, easily implementable, and is compatible with object-wise ground-truth depth. Specifically, we leverage intermediate features and responses for knowledge distillation. Considering long-range 3D dependencies, we propose \emph{3D-aware self-attention} and \emph{target-aware cross-attention} modules for student adaptation. Extensive experiments are performed to verify the effectiveness of our framework on the challenging KITTI 3D object detection benchmark. We implement our framework on three representative monocular detectors, and we achieve state-of-the-art performance with no additional inference computational cost relative to baseline models. Our code is available at https://github.com/rockywind/ADD.

\end{abstract}

\section{Introduction}
3D object detection is a primary foundation in 3D perception and has made evident contributions to autonomous driving, robot navigation and augmented reality.
To reach accurate 3D localization, some methods use LiDAR sensors \cite{shi2019pointrcnn, lang2019pointpillars, shi2020pv, he2020structure} or stereo cameras \cite{wang2019pseudo,li2019stereo,chen2020dsgn,sun2020disp} to provide accurate range measurements.
However, cost and complexity increases caused by using extra sensors result in limited generalization in downstream tasks. 
To solve these problems, the community has begun to pay more attention to monocular 3D object detection, which is much lower-cost and easily implementable.
%monocular 3D object detection methods \cite{li2020rtm3d,liu2021autoshape,liu2022learning,chen2021monorun,lian2022monojsg,ma2020rethinking,ding2020learning,shi2020distance,park2021dd3d,huang2022monodtr,wang2019pseudo,weng2019monocular,ma2019accurate,reading2021categorical}.

\begin{figure}
    \centering
    \includegraphics[width=8.5cm]{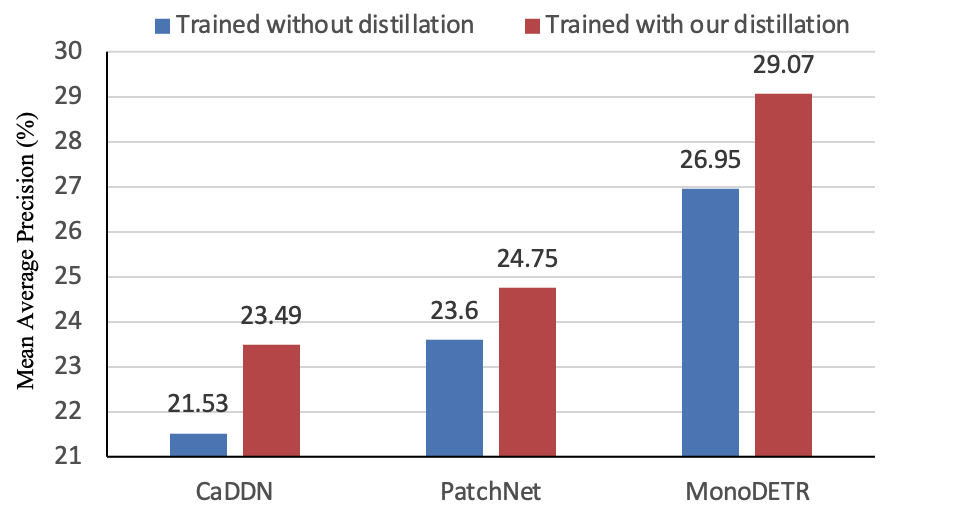}
    \caption{Moderate level BEV average precision ($\%$) on different baselines with (red bar) or without (blue bar) training the networks with our distillation framework on the KITTI validation split. Significant improvements are observed on three representative networks. More comparison results are presented in the Experiment Section.}
    \label{fig:fig1}
\end{figure}

To address the task of monocular 3D object detection, most deep-learning-based works adopt 2D-3D correspondences \cite{shi2021geometry,lu2021geometry,9578273,li2020rtm3d,liu2021autoshape,liu2022learning,chen2021monorun,lian2022monojsg} to constrain the learning of monocular images and directly estimate 3D bounding boxes.
However, because of the lack of explicit 3D measurements, their localization accuracy degenerates dramatically compared with LiDAR- or stereo-based methods.
To address this problem, some prior works~\cite{wang2019pseudo,ma2019accurate,ma2020rethinking} take monocular depth estimation networks~\cite{fu2018deep} to precompute depth estimations as extra inputs.
% Afterwards, they concatenate them to monocular inputs or use them as the depth clue to generate pseudo LiDAR point clouds.
Some other works~\cite{park2021dd3d,reading2021categorical,zhang2022monodetr}  take depth estimation as an auxiliary task to introduce depth-aware features for object detection.
However, inevitable depth estimation errors result in easily affected high-level features, thus leading to feature smearing and suboptimal predictions. 

% \begin{figure*}
%     \centering
%     \includegraphics[width=17cm]{Final-Figure2.png}
%     \caption{An overview of the proposed attention-based depth distillation framework pipeline.
%     % It consists of three main components: 
%     % (i) a depth-assisted monocular Student Network that takes depth GT for supervision; 
%     % (ii) an aligned Teacher Network with identical architecture as the student that takes depth GT as extra input, 
%     % and (iii) distillation branches between them. 
%     Firstly, we process and send object-/pixel-wise GT depth and monocular images into teacher network, whose architecture is identical to the student, and train it independently.
%     Secondly, we load and freeze teacher parameters, and leverage distillations on features and responses to supervise the student. 
%     Finally, we use  student network without teacher parameters for inference.}
%     \label{fig:fig2}
% \end{figure*}
To mitigate this issue, we propose an attention-based depth knowledge distillation (KD) framework with 3D-aware positional encoding (PE) to provide students with \textbf{mighty} and \textbf{dynamic} depth-positional cues.
% In detail, it consists of three main components: 
% (i) a depth-assisted monocular Student Network that takes GT depth for supervision; 
% (ii) an aligned Teacher Network with identical architecture as the student that takes GT depth as extra input, 
% and (iii) distillation branches between them. 
Unlike previous knowledge distillation frameworks that adopt stereo- or LiDAR-based teachers, we build up our teacher with identical architecture as depth-assisted student but with extra ground-truth (GT) depth as input.
Leveraging teachers based on GT depth helps provide students with extra regularization to alleviate detection overfitting caused by depth estimation errors.
Particularly, our framework is compatible with both object- and pixel-wise GT depth for teacher building up, i.e., not necessarily relying on extra LiDAR support.

Specifically, we conduct knowledge distillation on both intermediate features and network responses and leverage attention-based transformer modules to capture long-range 3D dependencies during student adaptation.
Firstly, we design a \emph{3D-aware self-attention module} to encode student intermediate features.
Motivated by using PE for depth and 3D embedding \cite{huang2022monodtr, liu2022petr, liu2022petrv2}, to involve stronger hints for 3D reasoning, we integrate semantic features with precise 3D PE from GT depth.
Credit to our KD-based framework, our PE in adaptation modules is discarded during inference, thus not bothering our framework's fairness relative to prior monocular detectors. 
Secondly, for encoder-decoder detecting heads~\cite{carion2020end}, we adopt a \emph{target-aware cross-attention module} for response adaptation, which revolves student object queries under the guidance of depth-aware teacher queries.

Compared with MonoDistill~\cite{chong2022monodistill}, whose teacher and student models are based on different data modalities, our framework is domain-gap free, as our teachers originate from monocular detectors and hence are easily implementable. 
Compared with Pseudo-stereo~\cite{chen2022pseudo}, which takes an advanced stereo-based network~\cite{guo2021liga} as a teacher, our teachers are easily accessible and effective as well.
Besides, architectural differences between stereo and monocular detectors result in potentially spatially unaligned features and responses, while our framework is seamless with guaranteed feature- and output-level consistency.

% \xz{阐述我们的知识蒸馏框架，包含3D aware attention based feature distillation以及target-aware attention based response distillation，并分别介绍每个distillation模块的基本思想。}
% \xz{然后再加上下面的两个compared with。这两个compared with写的很好。}

% \xz{另起一段，介绍我们把这个知识蒸馏框架用在了三个具有代表性的3D检测网络上，分别是XX，XX，和XX。图1展示了对应的结果，由此可见，我们的框架可以明显提升网络的检测性能。}

% \xz{Overall, our contributions can be summarized as follows:}
% \xz{\begin{item}逐点去介绍我们的贡献。至此，我们的intro就结束了。}
% \xz{你下面写的太细节了，怎么把蒸馏框架分别移植到那三个baseline网络上、怎么训练网络都不是intro应该介绍的，而是在实验部分写的。}

% We design our teachers with identical architectures as monocular students but with extra ground-truth (GT) depth as input.

Down to the aforementioned advantages, we successfully implement our framework on three representative monocular 3D object detection methods, including MonoDETR~\cite{zhang2022monodetr}, CaDDN~\cite{reading2021categorical} and PatchNet~\cite{ma2020rethinking} using the challenging KITTI 3D objection benchmark~\cite{geiger2012we}. The performance improvements are shown in Figure~\ref{fig:fig1}.
Overall, we summarize our contributions as follows.
\begin{itemize}
    \item[(i)] We propose a seamless KD framework for depth-assisted monocular 3D object detectors. 
    It injects mighty and dynamic depth-positional cues into students to boost their performances.
    Our teachers are easily accessible and effective with guaranteed feature- and output-level consistency.
    Our teacher-student framework is domain gap-free and hence easily implementable.
    \item[(ii)] We propose a 3D-aware self-attention module for the adaptation of multi-level intermediate features by integrating GT depth guided PE with student semantic features, as well as a target-aware cross-attention module for the adaptation of decoding head~\cite{carion2020end} responses.
    Our attention-based adaptation modules benefit from capturing long-range 3D dependencies.
    \item[(iii)] We conduct extensive experiments to show the effectiveness of our framework on three representative baselines using the challenging KITTI 3D object detection benchmark \cite{geiger2012we}. 
    Our model achieves state-of-the-art performance with no extra inference computational cost relative to the baselines.
    % Which makes it easily to implement compared with .
\end{itemize}

\begin{figure*}
    \centering
    \includegraphics[width=17cm]{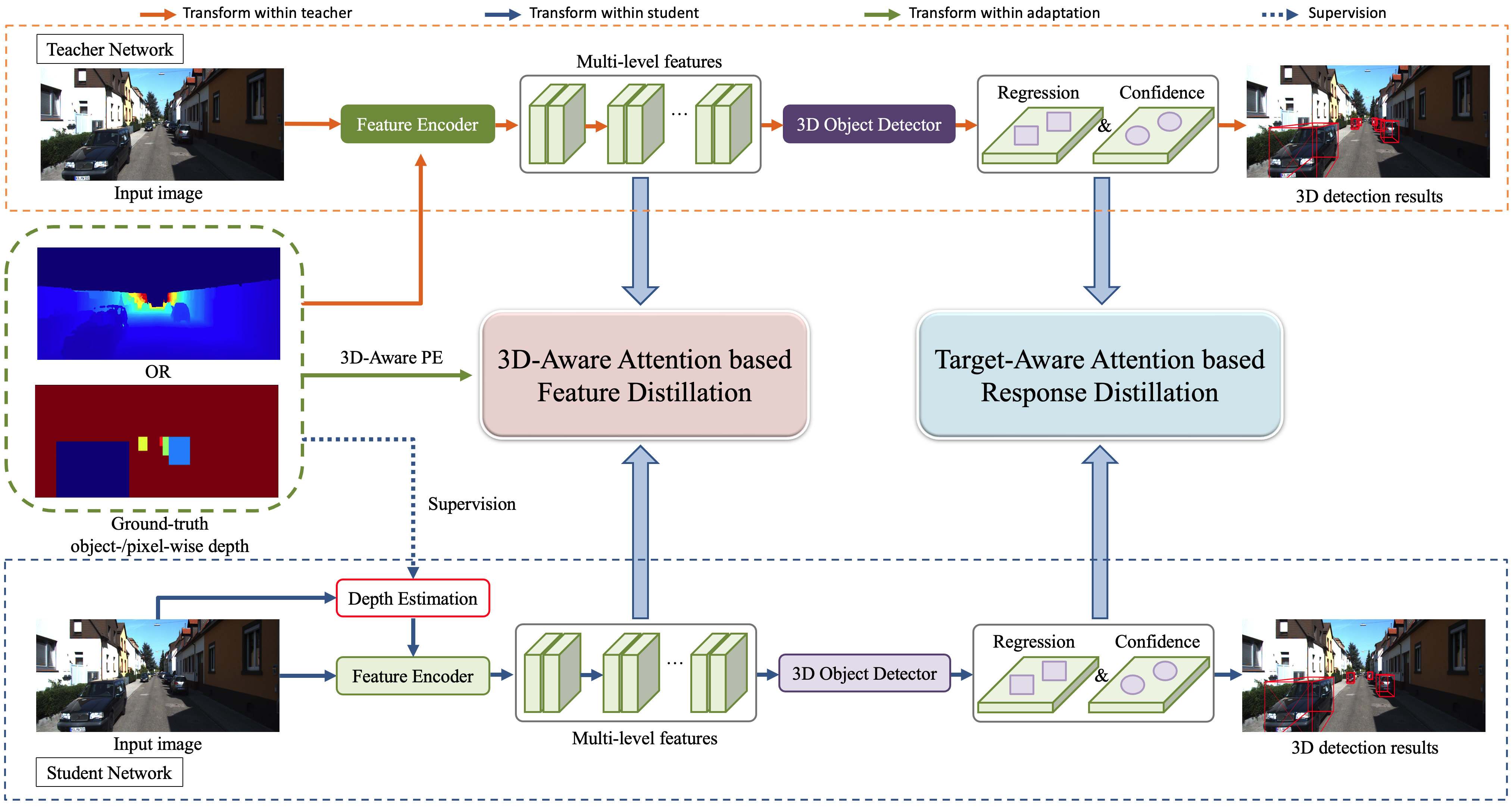}
    \caption{An overview of the proposed attention-based depth distillation framework pipeline.
    % It consists of three main components: 
    % (i) a depth-assisted monocular Student Network that takes depth GT for supervision; 
    % (ii) an aligned Teacher Network with identical architecture as the student that takes depth GT as extra input, 
    % and (iii) distillation branches between them. 
    Firstly, we process and send object-/pixel-wise GT depth and monocular images into teacher network, whose architecture is identical to the student, and train it independently.
    Secondly, we load and freeze teacher parameters, and leverage distillations on features and responses to supervise the student. 
    Finally, we use  student network without teacher parameters for inference.}
    \label{fig:fig2}
\end{figure*}

\section{Related Work}
\paragraph{Monocular 3D object detection.} \ 
Monocular 3D object detection takes as input only a single image, aiming at identifying objects of interest and localizing their 3D bounding boxes. 
To alleviate the ambiguity in 2D-3D projection caused by the lack of accurate 3D measurements, existing approaches either leverage neural networks to design semantic and geometric constraints or adopt external depth information for depth reasoning to facilitate 3D object detection.
For methods without depth assistance, Mousavian et al.~\cite{mousavian20173d} solved constraints between 2D bounding boxes and 3D dimensions to recover the 3D location. 
CenterNet \cite{zhou2019objects} proposed a centerness-based object detection paradigm that adds extra 3D task heads to lift the 2D detector to 3D space. 
% \xz{Why do you introduce CenterNet first? It seems that Mousavian's work was published earlier. It's better to introduce works according to the publication year.}
M3D-RPN~\cite{brazil2019m3d} proposed 3D anchors by aligning 2D anchors with 3D statistics.
Instead of directly regressing object-wise depth, some methods estimated 2D and 3D heights with uncertainty modeling to recover 3D locations based on geometric priors \cite{shi2021geometry,lu2021geometry}.
% MonoFlex \cite{9578273} pointed out that truncation leads to inconsistent offset distributions and handled it by disentangling the network with a sophisticated model ensemble.
Several works~\cite{li2020rtm3d,liu2021autoshape,liu2022learning, chen2021monorun} proposed keypoint-based approaches to limit the searching space of geometric constraints. 
% MonoRun \cite{chen2021monorun} proposed a self-supervised algorithm to learn pixel-level constraints.
Recently, MonoJSG \cite{lian2022monojsg} designed a cost volume from semantic and geometric cues to model the depth error in a two-stage manner.
However, an obvious drawback of the above methods is the decayed performance in 3D localization due to the lack of direct depth guidance.

On the other hand, for methods with depth assistance, some works~\cite{ma2020rethinking,ding2020learning,shi2020distance} accomplished monocular 3D object detection by directly taking estimated depth maps as extra inputs. 
DD3D \cite{park2021dd3d} leveraged GT depth maps for depth pretraining.
MonoDTR \cite{huang2022monodtr} generated intermediate pixelwise depth estimations to provide positional encoding to a depth-aware transformer. 
Other works~\cite{wang2019pseudo,weng2019monocular,ma2019accurate,reading2021categorical} jointly employed the estimated depth maps and intrinsic and extrinsic calibrations to project semantic features from monocular images into 3D space in a point cloud or voxel manner.
However, the performance of the above depth-assisted methods is greatly affected by the accuracy of the estimated depth.
Unfortunately, it is nontrivial to estimate an accurate depth from only 2D monocular images.

\paragraph{Knowledge distillation.} \ 
Knowledge distillation (KD) was first developed for model compression \cite{hinton2015distilling}, which makes student networks learn from both GT labels and soft labels generated from teacher networks. 
Romero et al.~\cite{romero2014fitnets} showed that student networks can also be guided by features from intermediate layers.
Afterwards, many methods successfully adopted knowledge distillation in various tasks, such as image classification \cite{heo2019comprehensive,tung2019similarity,yim2017gift,fukuda2017efficient}, semantic segmentation \cite{jiao2019geometry,liu2019structured} and depth estimation \cite{chen2021revealing,hu2021boosting}, etc. 

In the field of 2D object detection, Chen et al.~\cite{chen2017learning} first implemented knowledge distillation by distilling features from neck, classification head and regression head. 
To tackle the problem of foreground-background balancing, some works \cite{li2017mimicking,wang2019distilling} either mimic areas from the region proposal or distill the student with fine-grained features from foreground object regions. 
Guo et al. ~\cite{guo2021distilling} showed that distillation on regions excluding objects is also important.
In 3D object detection, LiGA-Stereo \cite{guo2021liga} guided the learning of a stereo-based detector by geometric-aware high-level features from a LiDAR-based detector to mitigate the influence of depth estimation error.
Recently, MonoDistill \cite{chong2022monodistill} designed a teacher based on inputs of a projected LiDAR signal to educate a monocular 3D detector with spatial cues.
Rather than using a LiDAR-based detector as a teacher, Chen et al.~\cite{chen2022pseudo} proposed distilling a monocular 3D detector from an advanced stereo 3D detector as there is a smaller domain gap in image-to-image than image-to-LiDAR.
Unlike previous works, our network comprises a series of novel distillation techniques on intermediate features and responses to transfer depth knowledge from the teacher network that is trained using GT depth maps to the student network, and greatly improves the 3D detection performance compared with the baselines without knowledge distillation.
In particular, our method takes no extra stereo or LiDAR data as input, thus expanding its application scope.

\section{Methodology}
As shown in Figure \ref{fig:fig2}, we design our attention-based depth distillation framework with three main components: 
(i) a depth-assisted monocular Student Network that takes GT depth for supervision, 
(ii) an aligned effective (see Table \ref{Teacher_Conparison}) Teacher Network with identical architecture as the student that takes GT depth as extra input, 
and (iii) distillation branches between them. 
We adopt knowledge distillation on both intermediate features and network responses, where we propose \emph{3D-aware self-attention} and \emph{target-aware cross-attention} for feature adaptation, respectively.

% \section{Preliminaries and Motivation}

\subsection{General Depth-assisted Model}
First, we briefly review the general pipeline for depth-assisted monocular 3D object detectors. 
Here, we denote $I_{M}$ as the monocular image input.
To accomplish object classification and 3D bounding box regression over all possible targets, the depth-assisted models estimate a corresponding depth map $\hat{D}_{M}$ from $I_{M}$ and integrate it with monocular semantic features. 
Formally, the model can be formulated as: 
\begin{equation}
\begin{aligned}
	\hat{D}_M &= G_{Dep}(I_M), \\
    F_{Cls}, F_{Box} &= G_D(G_E(T(I_M, \hat{D}_M))),
\end{aligned}
\end{equation}
where $G_{Dep}$ denotes a monocular depth estimation network that either predicts depth distributions or continuous depth values; $T$ denotes a data representation transformation that may project monocular features into 3D space or sustain them in a 2D manner; $G_E$ denotes a feature encoding architecture to fuse semantic and depth features; $G_D$ is a decoding head for generating target predictions from high-level features; $F_{Cls}$ and $F_{Box}$ are detection responses for classification and regression, respectively. 

Note that the accuracy of $\hat{D}_M$ directly affects the final object prediction performance, and different methods usually utilize different designs of $G_{Dep}$.
Specifically, some works~\cite{ma2019accurate,ding2020learning,ma2020rethinking} directly take estimations from pretrained monocular depth estimators as $\hat{D}_M$.  
CaDDN~\cite{reading2021categorical} jointly learns a depth distribution estimator and a 3D detector, and MonoDETR~\cite{zhang2022monodetr} learns depth-aware feature maps by using auxiliary depth supervision.
However, there is still a significant gap between the predicted and ground-truth depths.
Therefore, inevitable errors in depth priors result in easily affected high-level features, thus leading to sub-optimal predictions. 
To mitigate this problem, we propose a KD framework by using a teacher model equipped with precise 3D measurements to teach a student model for reliable depth estimation, thus facilitating accurate 3D object detection. 
Below, we shall discuss our considerations when designing such a KD-based framework.

\if 0
% \#2 existing distillation techniques + drawback
\paragraph{KD for Monocular 3D object detection.}
Pseudo-stereo~\cite{chen2022pseudo} achieves superior performance by adopting KD with a teacher of stereo-based 3D object detector. 
However, it requires an advanced stereo detector~\cite{guo2021liga} with careful architecture designs.
Besides, using a teacher with different underlying mechanisms leads to potential risk of feature- and object-level miss-alignment, i.e. not distillation friendly.
Our framework can resolve identified issues above by introducing a teacher with identical architectures as the student, but with precise depth GT.
Such a teacher design is effective but easily accessible.
It also ensures a general seamless KD framework with guaranteed feature- and object-level consistency.
MonoDistill~\cite{chong2022monodistill} leverages a teacher based on projected LiDAR inputs, i.e. pixel-wise depth maps, to guide a monocular detector merely based on monocular inputs. It leads to a huge teacher-student domain gap due to diverse data modalities, and relies on accurate pixel-level range measures from expensive LiDAR sensors.
Our framework instead is domain-gap free as both teacher and student models are originated from monocular detectors. 
In addition, our framework can achieve significant performance boost by only using teachers based on object-wise depth GT. 
\fi

\subsection{Considerations for Knowledge Distillation}
% We first observe that training depth assisted monocular detectors with corresponding depth GT significantly improves their detection performance, as shown in Table \ref{Teacher_Conparison}.
% It motivates us to leverage them for KD, because strong teachers inherently contribute student performance improvements. 
%Although depth-assisted detectors are always supervised with precise 3D measurements for once only, they show stronger performance than most direct monocular detectors.
For these depth-assisted detectors, we believe that they can show stronger localization ability if we could inject them with \textbf{mighty} and \textbf{dynamic} depth-positional cues.
Hence, intuitively, leveraging the KD mechanism with a depth-aware teacher is expected to be beneficial.
Before introducing our KD framework, we first discuss the relevant
design considerations that we have taken.
\begin{itemize}
\item[(i)]
Considering the difficulties and costs in 3D data collection, especially from erroneous multi-sensor alignment, we cannot introduce extra data requirements when designing teacher-student models.
For example, we should avoid using expensive LiDAR assistance when training the teacher model.
% An effective way of designing teachers for depth-assisted students is training them with corresponding depth GT, as shown in Table \ref{Teacher_Conparison}.
% \xz{不要写我们设计了什么，我们设计了什么是下一个subsection才介绍的，这里要写的是我们在设计的时候需要考虑什么因素。所以不应该是our deisgn enables XX，因为这个时候别人还不知道你的design是什么，而应该是when we design our network, we should consider XXX。所以(ii)-(v)的描述口吻应该转变一下。}
\item[(ii)]
Second, although introducing the dark knowledge from different domains benefits KD learning, it causes greater hardship for efficient implementation.
Thus, a desired framework needs to consider data distribution consistency between teacher and student data usages.
% Our framework is domain-gap free by leveraging teachers originated from monocular detectors.
\item[(iii)]
Third, to develop a seamless KD framework, we need to ensure both feature- and output-level consistency in the teacher-student model. 
Otherwise, it takes extra costs and complexity to deal with spatially-unaligned intermediate features or responses.
\item[(iv)]
Besides, it is usual to take teacher detectors with advanced designs for performance boosting.
However, a complex architecture is not flexible, which limits the generalization of such KD methods. 
Thus, we should use easily accessible architectures for teacher design.
\item[(v)]
To make effective feature approximation, it is important to adopt adaptation modules on student outcomes \cite{wang2019distilling} before taking distillation losses with teachers.
% \xz{Why feaature adaptation module is important?}
We believe that leveraging transformer feature adaptation modules based on attention mechanisms will contribute to 3D information reasoning as they could consider long-range 3D dependencies.
% \item[(v)]
% Although depth-assisted detectors are always supervised with precise 3D measurements for once only, they show stronger performance than most direct monocular detectors.
% We believe those detectors can show stronger localization ability if we could inject them with \textbf{mighty} and \textbf{dynamic} depth-positional cues.
% Hence, intuitively, using a depth-aware teacher is expected to be beneficial for meaningful 3D reasoning.
% It motivates us to leverage the KD mechanism as we could design a depth-aware teacher to generate multi-level features and responses for supervision monocular students.
\end{itemize}
% 1. 考虑到数据获取，尤其是三维数据采集和标注的困难，我们在设计网络时应该充分利用现有数据集提供的数据种类，而不能额外引入其他模态的数据，比如雷达点云。。。
% 2. 虽然引入额外的数据来源会丰富数据形式，但是也会引起domaim gap，增大蒸馏的难度，难以保证蒸馏效果。所以，我们在设计蒸馏网络时，要尽可能保证teacher-student这两个网络数据分布的一致性
% 3. 考虑到我们期望能够设计一个通用的知识蒸馏框架，所以当把我们的框架用在现有的深度辅助的网络上时，要具有灵活性（可移植性），能够以最小的结构改动就嵌入现有网络，带来最大的蒸馏效果

%\subsection{Overview of Our Method}

%We are going to show their details in the following subsections.
% \xz{这个overview写的不好。这一段其实就是看图说话，对照着图2，从左到右把整体流程介绍一遍。现在完全看不到任何对于图2的介绍。}

% \xz{overview介绍完了之后，剩下的subsection应该是：}
% \subsection{Self-Attention based Feature Distillation}
% \xz{这个子章节介绍粉色那个模块内部到底是怎么设计的}
% \subsection{Cross-Attention based Response Distillation}
% \xz{这个子章节介绍蓝色那个模块内部到底是怎么设计的}
% \subsection{End-to-end Network Training}
% \xz{这个子章节介绍网络训练用的损失函数}

\begin{figure}
\centering
\includegraphics[width=8.5cm]{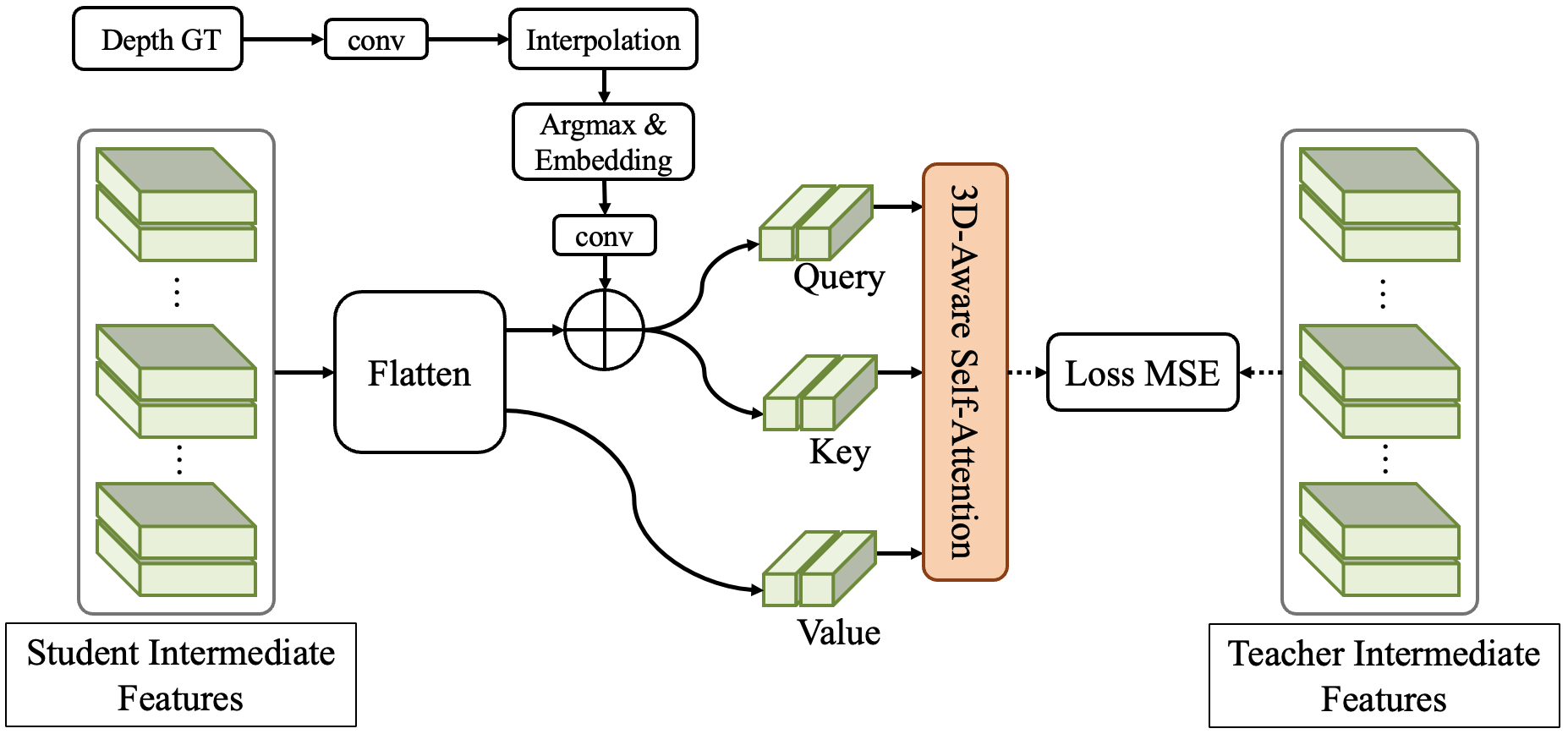}
\caption{The visualization of our KD on multi-level intermediate features with 3D-aware positional encoding for feature adaptation.}
\label{fig:fig3}
\end{figure}

% \begin{figure}
% \centering
% \includegraphics[width=8.5cm]{Final-Figure4.png}
% \caption{The visualization of our KD on responses with target-aware cross-attention for feature adaptation.}
% \label{fig:fig4}
% \end{figure}

\subsection{3D-aware Attention based Feature Distillation}
% \paragraph{Distillation via Intermediate Features.}
Our feature-level distillation pipeline is shown in Figure \ref{fig:fig3}. We define different levels of student and teacher intermediate features obtained from convolutional neural network (CNN) backbones as $\{F^k\}_{k\in(1,...,n)}$ and $\{T^k\}_{k\in(1,...,n)}$ respectively, where $n$ is the total number of intermediate feature levels.
Besides, we denote the depth GT as $\mathcal{D}$.
It is important to consider the information imbalance between foreground and background regions when adopting KD on detection tasks. 
We integrate features from these two types of regions by introducing binary masks $M$ based on 2D bounding boxes $B^k$:
\begin{equation}
   M^k=\{M^k_{i,j}|M^k_{i,j} = \mathbb{I}\left[(i,j)\in B^k\right]\},
\end{equation}
where $i,j$ varies over the width and length of $F^k$, and $M^k_{i,j}$ equals one if position $i,j$ of feature map $F^k$ belongs to the foreground but zero otherwise.

\paragraph{3D-aware position encoding.}
Inspired by the leverage of positional encoding (PE) for depth and 3D embedding \cite{huang2022monodtr, liu2022petr, liu2022petrv2} with estimated depth and discretized meshgrid coordinates, credit to our KD-based framework, we are able to introduce semantic features with more accurate 3D positional encoding under the guidance of depth GT. 
Specifically we generate 3D-aware positional encodings $F_{emb}$ under the guidance of $\mathcal{D}$, and introduce $F_{emb}$ to self-attention operations on student features.
We thus adopt a multi-layer perceptron (MLP) and a bilinear interpolation on $\mathcal{D}$ to produce encoded depth feature $F_d$, which has $C_d$ channels and match the size of $F^k$ .
Afterwards, we solve its argmax value at the channel dimension, denoted as $G_{arg}(\cdot)$, to obtain a map of the depth positional index. 
We send positional index maps to a position embedding $G_{pe}(\cdot)$ with dictionary size $C_d$ and embedding dimension $C_{dim}$ to produce the 3D embedding maps $F_{emb}$.
Hence we send $F_{emd}$ to another MLP, flatten output embedding maps into vectors, and point-wisely add them to flattened $F^k$ to generate semantic queries and keys with 3D-aware positional encoding. 
In function, we have:
\begin{equation}
    F_{emb} = G_{pe}(G_{arg}(G_A(\mathcal{D}))),
\end{equation}
where $G_A$ represents implementing an MLP and a bilinear interpolation. 
To ease the computational burden, we build up our MLP with convolution layers of kernel size 1.

Afterwards, queries and keys interact with semantic values in a self-attention manner to generate 3D-aware semantic features $F^k_{3d}$. 
Taking in our foreground masks, we have the distillation loss for intermediate features as :
\begin{equation}
    \mathcal{L}_{feat} =\sum_{k=1}^n \alpha_I \|M^k(F^k_{3d}-T^k)\|^2 + \beta_I \|(1-M^k)(F^k_{3d}-T^k)\|^2,
\end{equation}
where $\alpha_I$ and $\beta_I$ are hyperparameters for feature-level foreground and background balancing.

% \subsection{XXX cross-attention module}
\subsection{Target-aware Attention based Response Distillation}
\begin{figure}
\centering
\includegraphics[width=8.5cm]{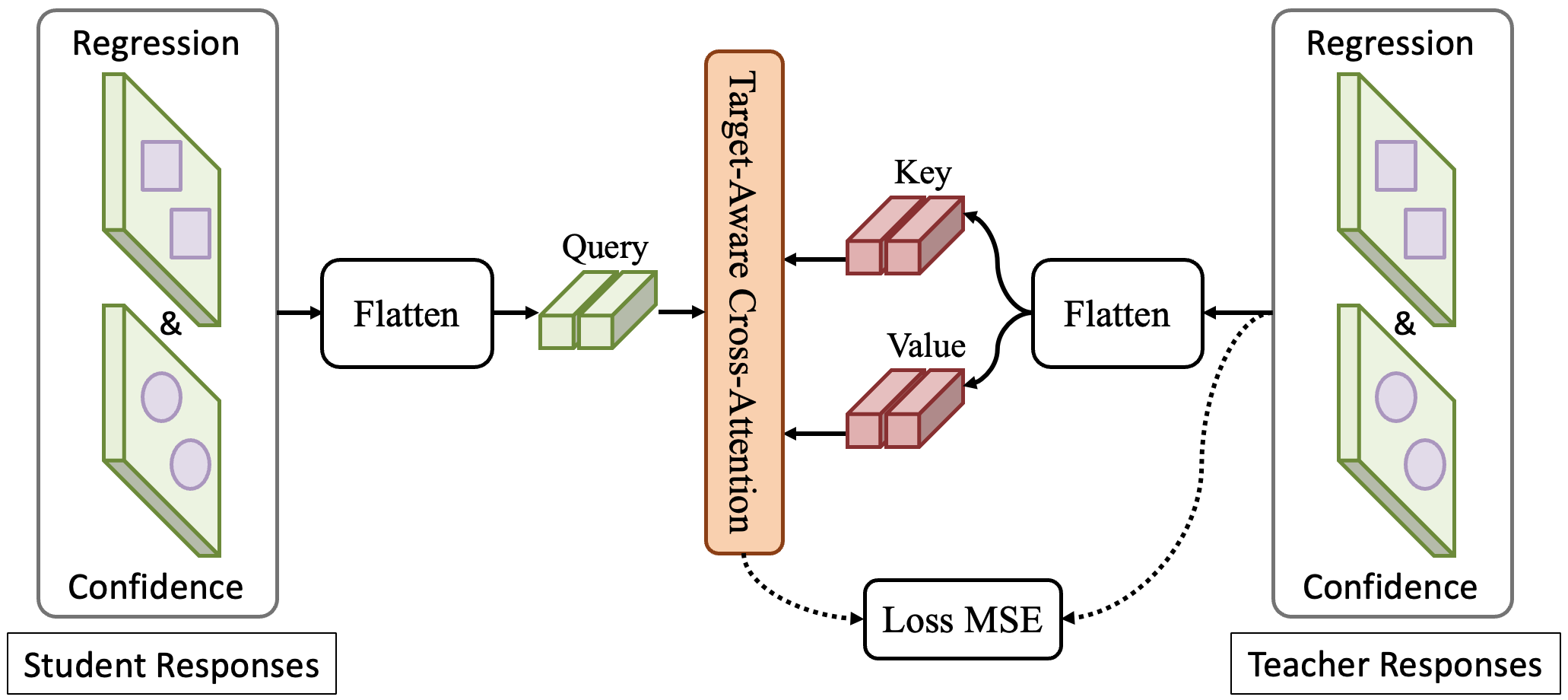}
\caption{The visualization of our KD on responses with target-aware cross-attention for feature adaptation.}
\label{fig:fig4}
\end{figure}

Our output-level distillation pipeline is shown in Figure \ref{fig:fig4}.
For detectors with an encoder-decoder transformer structure \cite{carion2020end} for object detection, we define student and teacher decoders' intermediate object queries as $\{F^k_v\}_{k\in(1,...,m)}$ and $\{T^k_v\}_{k\in(1,...,m)}$, where $m$ is the number of decoders' repeated transformer blocks. 
To achieve proper foreground and background balancing, we adopt the efficient Hungarian algorithm \cite{stewart2016end} between teacher predictions and hard labels to generate a foreground query mask $M_f$ for all object query levels. 
We leverage a cross-attention design as our feature adaptation module to guide student queries to learn meaningful 3D reasoning from the teacher.
Specifically, we adopt MLPs to produce $q^k_v$ from $F^k_v$, and produce $k^k_v$,$v^k_v$ from $T^k_v$.
Afterwards, we integrate these vectors in a cross-attention manner with $q^k_v$ as the query, $k^k_v$ as the key, and $v^k_v$ as the value to generate adapted student object queries $F^k_a$.
We have our distillation loss as:
\begin{equation}
    \mathcal{L}_{ed} = \sum_{k=1}^m \alpha_v\|M_f(F^k_a-T^k_v)\|^2 + \beta_v\|(1-M_f)(F^k_a-T^k_v)\|^2,
\end{equation}
where $\alpha_v$ and $\beta_v$ are hyperparameters for response-level foreground and background balancing.

\subsection{End-to-end Network Training}
We denote inherited losses from depth-assisted students as $\mathcal{L}_{cls}, \mathcal{L}_{reg}$ and $\mathcal{L}_{depth}$ for object classification, 3D bounding box regression and depth regression, respectively. 
$\mathcal{L}_{depth}=0$ if baseline models utilize estimations from pretrained depth estimators. 
Our total loss function is:
\begin{equation}
    \mathcal{L} = \mathcal{L}_{cls} + \mathcal{L}_{reg} + \mathcal{L}_{depth} + \alpha\mathcal{L}_{feat} + \beta\mathcal{L}_{ed},
\end{equation}
where $\alpha$ and $\beta$ are hyperparameters for loss balancing.
% *************************************AP BEV*******************************************
\begin{table*}[t]
\centering
\resizebox{\textwidth}{22mm}{
\begin{tabular}{c|ccc|ccc|ccc}
\hline
\multirow{2}{*}{Method}
     & \multicolumn{3}{c|}{$AP_{BEV}$@IoU=0.7(Car test)}
     & \multicolumn{3}{c|}{$AP_{BEV}$@IoU=0.7(Car val)} 
     & \multicolumn{3}{c}{$AP_{3D}$@IoU=0.7(Car val)} \\  \cline{2-10}
     & Easy & Mod. & Hard & Easy & Mod. & Hard & Easy & Mod. & Hard  \\ \hline\hline
    MonoDETR~\cite{zhang2022monodetr} & 33.60 & 22.11 & 18.60 & 37.86 & 26.95 & 22.80& 28.84 & 20.61 & 16.38 \\
    MonoDETR$\mathbf{+ADD}$ & 35.20 & 23.58 & 20.08 & 40.38 & 29.07 & 25.05  & 30.71 & 21.94 & 18.42 \\
    Improvement  & \textcolor{red}{+1.60} & \textcolor{red}{+1.47} & \textcolor{red}{+1.48}& \textcolor{red}{+2.52} & \textcolor{red}{+2.12} & \textcolor{red}{+2.25}  & \textcolor{red}{+1.87} & \textcolor{red}{+1.33} & \textcolor{red}{+2.04} \\\hline
    CaDDN~\cite{reading2021categorical} & 27.94 & 18.91 & 17.19& 30.28 & 21.53 & 18.96 & 23.57 & 16.31 & 13.84\\ 
    CaDDN$\mathbf{+ADD}$& 30.11 & 20.80 & 18.04 & 34.14  & 23.49 & 21.24  &  25.30 & 16.64 & 14.90\\
    Improvement & \textcolor{red}{+2.17} & \textcolor{red}{+1.89} & \textcolor{red}{+0.85}  & \textcolor{red}{+3.86} & \textcolor{red}{+1.96} & \textcolor{red}{+2.28} & \textcolor{red}{+1.73} & \textcolor{red}{+0.33} & \textcolor{red}{+1.06} \\\hline
    PatchNet~\cite{ma2020rethinking} & 22.97 & 16.86 & 14.97& 41.49 & 23.60 & 19.93 & 31.60 & 16.80 & 13.80\\
    % PatchNet~\cite{ma2020rethinking} & 41.49 & 23.60 & 19.93 & 26.34 & 16.65 & 13.81 \\\hline
    PatchNet$\mathbf{+ADD}$ & 28.15& 17.38 & 15.06& 42.15 & 24.75 & 20.26 &  32.21 & 16.92 & 13.87 \\
    Improvement & \textcolor{red}{+5.18} & \textcolor{red}{+0.52} & \textcolor{red}{+0.09}  & \textcolor{red}{+0.66} & \textcolor{red}{+1.15} & \textcolor{red}{+0.33} & \textcolor{red}{+0.61} & \textcolor{red}{+0.12} & \textcolor{red}{+0.07}\\ \hline
\end{tabular}
}
\caption{Quantitative comparisons of the Car category on the KITTI validation and testing splits. The results are evaluated using 40 recall positions.
Our improvements relative to baseline models are listed in \textcolor{red}{red}. }
% $^*$We use our retrained PatchNet as baseline for comparison because it provides no test set 2d predictions for data preparing.}
\label{BEV_Comparison}
\end{table*}
\begin{table*}[t]
\centering
\setlength{\tabcolsep}{0.75mm}{
\begin{tabular}{c|c|c|ccc|ccc}
\hline
\multirow{2}{*}{Methods}
         & \multirow{2}{*}{Reference}
    & \multirow{2}{*}{Extra Data}
      & \multicolumn{3}{c}{$AP_{BEV}$(Car test)}
      & \multicolumn{3}{c}{$AP_{3D}$(Car test)}\\  \cline{4-9}
      & & & Easy & Mod. & Hard & Easy & Mod. & Hard \\ \hline\hline
    PatchNet~\cite{ma2020rethinking} & ECCV 2020 & LiDAR & 22.97 & 16.86& 14.97 & 15.68 & 11.12 & 10.17 \\
    D4LCN~\cite{ding2020learning}& CVPR 2020& LiDAR & 22.51 &16.02 &12.55 & 16.65 & 11.72 & 9.51 \\
    DDMP-3D~\cite{wang2021depth}& CVPR 2021 & LiDAR &28.08 &17.89 &13.44& 19.71 & 12.78 & 9.80\\
    CaDDN~\cite{reading2021categorical}& CVPR 2021& LiDAR& 27.94 &18.91 &17.19& 19.17 & 13.41 & 11.46\\
    MonoDTR~\cite{huang2022monodtr}& CVPR 2022& LiDAR &28.59 &20.38 &17.14& 21.99 & 15.39 & 12.73\\
    Kinematic3D~\cite{brazil2020kinematic} & ECCV 2020& Temporal &26.69 &17.52 &13.10& 19.07 & 12.72 & 9.17\\\hline
    MonoDLE~\cite{ma2021delving}& CVPR 2021& None &24.79 &18.89 &16.00& 17.23 & 12.26 & 10.29 \\
    MonoRUn~\cite{chen2021monorun}& CVPR 2021 & None &27.94 &17.34 &15.24& 19.65 & 12.30 & 10.58 \\
    GrooMed-NMS~\cite{kumar2021groomed}&CVPR 2021&None &26.19 &18.27 &14.05& 18.10 & 12.32 & 9.65\\
    MonoRCNN~\cite{shi2021geometry}&ICCV 2021& None&25.48 &18.11 &14.10& 18.36 & 12.65 & 10.03\\
    MonoEF ~\cite{zhou2021monocular}&CVPR 2021&None &29.03 &19.70 &17.26& 21.29 & 13.87 & 11.71\\
    MonoFlex~\cite{9578273}& CVPR 2021&None&28.23 &19.75 &16.89& 19.94 & 12.89 & 12.07\\
    
    MonoJSG~\cite{lian2022monojsg}& CVPR 2022& None &$\mathbf{32.59}$ &21.26 &18.18& $\mathbf{24.69}$ & 16.14 & 13.64\\
    MonoCon~\cite{liu2022learning}& AAAI 2022& None &31.12 & $\mathbf{22.10}$ &  $\mathbf{19.00}$& 22.50 & $\mathbf{16.46}$ & \textcolor{blue}{13.95} \\\hline
    $\mathbf{Ours}$ & - & None &\textcolor{blue}{35.20} &\textcolor{blue}{23.58} &\textcolor{blue}{20.08}  & \textcolor{blue}{25.61}    &\textcolor{blue}{16.81}  & $\mathbf{13.79}$\\
    \multirow{3}{*}{Improvement} & - & \emph{v.s. LiDAR} &\textcolor{red}{+6.61} & \textcolor{red}{+3.2} & \textcolor{red}{+2.89}&\textcolor{red}{+3.62} & \textcolor{red}{+1.42} & \textcolor{red}{+1.06}\\
     & - & \emph{v.s. Temporal} &\textcolor{red}{+8.51} & \textcolor{red}{+6.06} & \textcolor{red}{+2.89}&\textcolor{red}{+6.54} & \textcolor{red}{+4.09} & \textcolor{red}{+4.62}\\
     & - & \emph{v.s. None} &\textcolor{red}{+2.61} & \textcolor{red}{+1.48} & \textcolor{red}{+1.08}&\textcolor{red}{+0.92} & \textcolor{red}{+0.35} & \textcolor{red}{-0.16}\\\hline

\end{tabular}
}
\caption{Quantitative comparisons of the Car category on the KITTI test split. The best results are listed in \textcolor{blue}{blue} and the second in $\mathbf{bold}$. Our improvements relative to the best of each `Extra Data' category are listed in \textcolor{red}{red}.}
\label{test_compare}
\end{table*}

\section{Experiments}
\paragraph{Dataset.}
We evaluate our framework on the challenging KITTI~\cite{geiger2012we} 3D object detection benchmark.
Following 3DOP~\cite{chen20153d}, we separate the dataset into 3,712 training samples, 3,769 validation samples, and 7,518 test samples.
%7,481 image samples for training and 7,518 image samples for testing, with synchronized LiDAR point clouds.
%Following  \cite{chen20153d}, it turns out to be a common practice to divide training samples into a training split with 3,712 samples and a validation split with 3,769 samples.
Ablation study was evaluated on the validation split with models trained on the training split.
There are three object classes of interest in the KITTI dataset: car, pedestrian and cyclist. 
Each class is divided into three difficulty levels (easy, moderate and hard) based on occlusion, truncation and size. 
% % ***********************************Ablation on Test_cp*********************************************
% \begin{table}[t]
% \centering
% % \setlength{\tabcolsep}{0.75mm}{
% % \setlength{\tabcolsep}{0.75mm}{
% % \resizebox{\textwidth}{15mm}{
% \begin{tabular}{c|ccc}
% \hline
% \multirow{2}{*}{Settings}
%      & \multicolumn{3}{c}{$AP_{3D}$@IoU=0.7(Car val)}\\  \cline{2-4}
%      & Easy & Mod. & Hard \\ \hline\hline
%     Baseline & 27.63 & 19.93 & 16.70\\\hline
% \end{tabular}
% % }
% \caption{

% }
% \label{Test_cp}
% \end{table}
% % ******************************************************************************** 

% ***********************************Teacher V Student*********************************************
\begin{table}[t]
\centering
\setlength{\tabcolsep}{0.5mm}{
\begin{tabular}{c|c|ccc|ccc}
\hline
\multirow{2}{*}{Method} & \multirow{2}{*}{Depth}
     & \multicolumn{3}{c|}{$AP_{BEV}$(Car val)} 
     & \multicolumn{3}{c}{$AP_{3D}$(Car val)}\\  \cline{3-8}
     & & Easy & Mod. & Hard & Easy & Mod. & Hard \\ \hline\hline
    \multirow{2}{*}{MonoDETR} & - & 37.86 & 26.95 & 22.80 & 28.84 & 20.61 & 16.38\\
    & \checkmark & 51.99 & 36.44 & 30.33 & 41.43 & 28.15 & 22.84 \\ \hline
    \multirow{2}{*}{CaDDN} & - & 30.28 & 21.53 & 18.96& 23.57 & 16.31 & 13.84 \\ 
    & \checkmark & 86.65 & 70.92 & 66.16 & 72.77 & 53.16 & 48.97\\ \hline
    \multirow{2}{*}{PatchNet} & - & 41.43 & 24.66 & 20.13 & 31.78 & 16.60 & 13.61\\
    & \checkmark & 88.05 & 72.06 & 60.64 & 78.55 & 55.13 & 46.29\\\hline
\end{tabular}
}
\caption{Quantitative comparisons between our teachers and baselines on the KITTI validation split. `Depth' refers to using the corresponding depth GT for training and inference.}
\label{Teacher_Conparison}
\end{table}
% ********************************************************************************

\paragraph{Evaluation metric.}
Following prior works \cite{guo2021liga, reading2021categorical, huang2022monodtr}, we adopt two vital evaluation metrics, BEV Average Precision ($AP_{BEV}$) and 3D Average Precision ($AP_{3D}$), to analyze our framework.
They are calculated using class-specific thresholds with 40 recall positions based on the intersection-over-union (IoU) of 2D BEV and 3D bounding boxes, respectively.
% Following prior works \cite{guo2021liga, reading2021categorical, huang2022monodtr}, our framework performance is measured by an IOU-based criteria to compute mean average precision .
Specifically, the car category has an IoU threshold of 0.7 while the pedestrian and cyclist categories have an IoU threshold of 0.5.

% \xz{缺少一个paragraph去介绍我们的三个baseline方法，否则implementation details那里提到baseline就会很突兀。}

\paragraph{Baseline models.}
To show the effectiveness of our KD framework, we implement it on three recent depth-assisted monocular detectors, MonoDETR~\cite{chong2022monodistill}, CaDDN~\cite{reading2021categorical} and PatchNet~\cite{ma2020rethinking}. 
Particularly, MonoDETR~\cite{chong2022monodistill} leverages an encoder-decoder transformer structure \cite{carion2020end} for object detection.
% while CaDDN~\cite{reading2021categorical} and PatchNet~\cite{ma2020rethinking} adopt anchor-based decoding heads.

\paragraph{Implementation details.}
Our framework is trained on NVIDIA V100 GPUs, with $\alpha_I=1.0$, $\beta_I = 0.1$ for feature-level balancing, $\alpha_v=1.0$, $\beta_v=0.5$ for response-level balancing and $\alpha=1.0$, $\beta=1.0$ for final loss balancing. 
We set $\beta=0$ if a baseline model does not adopt the transformer based encoder-decoder detecting head \cite{carion2020end}. 
Settings for optimizers, batch sizes and numbers of GPUs used follow those for baseline models~\cite{ma2020rethinking, reading2021categorical, zhang2022monodetr}.
We design our framework teachers based on corresponding depth GT used in students. 
Specifically, we use object-wise depth for MonoDETR~\cite{zhang2022monodetr} teacher, and pixel-wise depth for CaDDN~\cite{reading2021categorical} and PatchNet~\cite{ma2020rethinking} teachers.
We make the usage of depth GT following baselines, and report our teacher performance in Table \ref{Teacher_Conparison}.

% *******************************Ablation Study*************************************************
\begin{table*}[t]
\centering
\begin{tabular}{c|c|cccc|ccc|ccc}
\hline
     & \multirow{2}{*}{\#}
     & \multirow{2}{*}{F}
     & \multirow{2}{*}{$+SA$}
     & \multirow{2}{*}{R}
     &\multirow{2}{*}{$+CA$}
     & \multicolumn{3}{c|}{$AP_{3D}$@IoU=0.7(Car val)}
     & \multicolumn{3}{c}{$AP_{BEV}$@IoU=0.7(Car val)}\\  \cline{7-12}
     & & & & & & Easy & Mod. & Hard & Easy & Mod. & Hard \\ \hline
    \multicolumn{6}{c|}{Teacher(with object-wise depth GT)} & 41.43 & 28.15 & 22.84 & 51.99 & 36.44 & 30.33\\\hline\hline
    \multicolumn{6}{c|}{MonoDETR$^*$\cite{zhang2022monodetr}} & 27.63 & 19.93 & 16.70& 36.64 & 26.32 & 22.35 \\ \hline
    \multirow{2}{*}{Feature-level}&(a) & \checkmark &- &- &- & 28.92 & 19.86 & 17.21 & 38.08 & 26.71 & 22.84\\ 
    &(b) & \checkmark &\checkmark &- &- &29.50 &20.90 & 17.51 & 38.62 & 27.03 & 23.06\\\hline
    \multirow{2}{*}{Object-level}&(c) & - &- &\checkmark & - &29.00 & 19.97 & 16.58 & 37.93 & 26.66 & 22.66 \\
    &(d) & - &- &\checkmark &\checkmark 
    & 29.41 & 21.18 & 17.77& 38.69 & 27.27 & 23.32\\\hline
    &(e) & \checkmark &- &\checkmark &- 
    & 27.40 & 19.73 & 16.37 & 36.94 & 26.68 & 21.89\\
    Feature-level $\&$ &(f)& \checkmark &\checkmark &\checkmark &- 
    & 28.51& 20.63 & 17.28 & 38.02 & 26.81 & 22.96\\
    Object-level&(g)& \checkmark &- &\checkmark &\checkmark 
    & 29.20 & 20.69 & 17.25 & 37.95 & 26.63 & 22.73\\
    &(h)& \checkmark &\checkmark &\checkmark &\checkmark 
    & 30.71 & 21.94 & 18.42 & 40.38 & 29.07 & 25.05\\\hline
    \multicolumn{6}{c|}{Improvement} & \textcolor{red}{+3.08} & \textcolor{red}{+2.01} & \textcolor{red}{+1.72} & \textcolor{red}{+3.74} & \textcolor{red}{+2.75} & \textcolor{red}{+2.70}\\\hline
\end{tabular}
% }
\caption{
Ablation studies on the KITTI validation split. 
F denotes distillation on multi-level intermediate features, and $+SA$ denotes adopting our 3D-aware self-attention module for feature-level adaptation.
R denotes distillation on network responses, and $+CA$ denotes adopting our target-aware cross-attention module for response-level adaptation.
Our improvements are listed in \textcolor{red}{red}.
$^*$The results for MonoDETR~\cite{huang2022monodtr} are obtained by training the officially publicized code with our computational environment.}
\label{Ablation_Study}
\end{table*}
% ********************************************************************************

% ***********************************Ablation on SA*********************************************
\begin{table}[t]
\centering
\begin{tabular}{c|ccc}
\hline
\multirow{2}{*}{Settings}
     & \multicolumn{3}{c}{$AP_{3D}$@IoU=0.7(Car val)}\\  \cline{2-4}
     & Easy & Mod. & Hard \\ \hline\hline
    Baseline & 27.63 & 19.93 & 16.70\\\hline
    $+ SA$ w/o PE & 28.45 & 20.37 & 16.30  \\
    $+ SA$ w/ estimated PE & 27.99 & 20.19 & 17.01  \\ 
    $+ SA$ w/ GT PE &29.50 &20.90 & 17.51 \\\hline
    $\alpha_I=1.00, \beta_I=0.00$ & 28.37 & 20.32 & 17.12 \\
    $\alpha_I=1.00, \beta_I=0.05$ & 28.66 & 20.53 & 16.52 \\
    $\alpha_I=1.00, \beta_I=0.10$ &29.50 &20.90 & 17.51 \\
    $\alpha_I=1.00, \beta_I=0.20$ & 28.51 & 20.45 & 17.24 \\\hline
    % $\alpha_I=1.00, \beta_I=0.50$ & 29.24 & 20.24 & 16.87 \\\hline
    $\alpha_v=1.00, \beta_v=0.10$ & 28.55 & 20.04 & 16.65 \\
    $\alpha_v=1.00, \beta_v=0.25$ & 29.27 & 20.18 & 17.54 \\
    $\alpha_v=1.00, \beta_v=0.50$  & 29.41 & 21.18 & 17.77\\
    $\alpha_v=1.00, \beta_v=1.00$ & 28.85 & 20.65 & 17.32 \\\hline
\end{tabular}
% }
\caption{
Ablation studies for feature and response distillation modules on the KITTI validation split.
`estimated PE' denotes using the student's depth estimation sub-network to generate positional encoding.
`GT PE' denotes using object-wise depth GT to generate positional encoding.}
\label{AB_SA}
\end{table}
% ******************************************************************************** 
\subsection{Evaluation of Our Framework}
\paragraph{Comparisons with baselines.}
To evaluate the effectiveness of our framework, we implemented it on three depth-assisted baselines and re-trained each network model on KITTI 3D object detection dataset. 
The detection results of the Car category are reported in Table \ref{BEV_Comparison}.
For easy observation, the percentages of AP improvements are shown in red.
Clearly, equipped with our attention-based depth distillation framework, the performance of all three baselines is further improved significantly on both validation and test samples, showing that our framework can successfully transfer precise depth knowledge from teacher model to student.

While performance gains on PatchNet~\cite{ma2020rethinking} are not as obvious as those for MonoDETR~\cite{zhang2022monodetr} and CaDDN~\cite{reading2021categorical} on the validation split.
According to Simonelli et al.~\cite{simonelli2021we}, the main reason for this is the contamination of objection detection validation split~\cite{chen20153d} by depth estimation training split.
It leads to that partial depth maps (1226/3769) used for detection validation, obtained from DORN estimation~\cite{fu2018deep}, are as competitive as GT depth from direct LiDAR projection, and thus retrains the effectiveness of guiding monocular student with precise 3D measurements.

\paragraph{Comparisons with state-of-the-art methods.}
In Table \ref{test_compare}, we present the experimental results of our framework and other state-of-the-art methods on the KITTI test split.
Our framework achieves the best performance of the car class on three BEV-level metrics.
Compared with the second-best approaches, our framework outperforms them with the ratio of $8.01\%$, $6.70\%$ and $5.68\%$ on the easy, moderate and hard levels, respectively.
Furthermore, our $\mathbf{ADD}$ framework does not introduce any extra data or computational cost in the inference stage, hence is industrially implementable.

\subsection{Ablation Study}
In this section, we investigate the effects of each component of our framework on the KITTI validation split.

\paragraph{Feature and response distillation.}
% The results are reported in Table \ref{Ablation_Study}.
% It illustrates the effectiveness of our 3D-aware self-attention on feature-level KD and target-aware cross-attention on output-level KD.
We evaluate the effectiveness of our 3D-aware self-attention on feature-level KD and target-aware cross-attention on output-level KD in Table \ref{Ablation_Study}.
The first row Teacher(with object-wise depth GT) is the performance of our monocular teacher based on object-wise depth GT.
The second row is our reproduce of MonoDETR\cite{zhang2022monodetr}, but with slightly decayed performance compared with the official declaration due to environmental differences.
Our framework enhances the student performance by large margins on all six metrics.

\paragraph{Settings for PE and loss weights.}
We first study whether our proposed  3D-aware positional encoding helps feature adaptation. 
From Table \ref{AB_SA} we observe that precise 3D measurements from GT PE can provide the most effective spatial guidance.
We then study the influence of feature-level foreground-background balancing, which illustrates that limiting the background loss weight to one tenth of the foreground achieves the best performance.
Afterwards, considering response-level foreground-background balancing, Table \ref{AB_SA} shows that the optimal choice is setting the background loss weight as half of the foreground.

\section{Conclusion}
% \section{Conclusion and Limitation}
In this work, we present $\mathbf{ADD}$, an attention-based depth distillation framework for monocular 3D object detectors.
We propose a \emph{3D-aware self-attention module} for intermediate feature distillation and a \emph{target-aware cross-attention module} for response distillation.
Our framework does not necessarily rely on LiDAR or stereo assistance.
In practice, our framework provides students with extra regularization to alleviate detection overfitting caused by depth estimation errors.
At a high level, our framework injects students with mighty and dynamic depth-positional cues for 3D reasoning.
We successfully implement our framework on three representative baselines using the KITTI 3D object detection benchmark, and achieve state-of-the-art performance without extra computational cost during inference.

{\small

\bibliography{aaai23}
}

\end{document}